\documentclass[journal,twoside]{IEEEtran}

\usepackage{cite}
\usepackage{microtype}

\ifCLASSINFOpdf
  \usepackage[pdftex]{graphicx}
\else
  \usepackage[dvips]{graphicx}
\fi
\usepackage{placeins}

\usepackage{amsmath}
\usepackage{amssymb}  
\usepackage{mathtools}

\usepackage{algorithm} 
\usepackage{algpseudocode} 
\usepackage{array}
\usepackage[caption=false,font=footnotesize]{subfig}
\usepackage{url}
\usepackage{siunitx}
\usepackage{hyperref}
\usepackage{multirow}
\usepackage{threeparttable}
\usepackage[multiple]{footmisc}

\usepackage{color}
\usepackage[dvipsnames]{xcolor}
\usepackage{verbatim}

\usepackage{xcolor} 
\usepackage[normalem]{ulem}

\begin{document}
\title{PolyMorph: Learning Locomotion Policy for Cross-Embodiment Adaptation}

\author{Hyunsik Oh, Seunghun Jeon, Donghoon Youm, Jungwoo Hur, Hyeongjun Kim, \\Jeong Hyun Lee, Moonkyu Jung, and Jemin Hwangbo$*$ 
}


\maketitle

\begin{abstract}
Reinforcement learning has enabled robust locomotion for legged robots, but learned policies are typically tied to a specific robot morphology and require additional engineering when the number of legs, actuators, or mechanical properties change. We present PolyMorph, a morphology-conditioned framework that learns a single locomotion policy for heterogeneous legged robots. PolyMorph converts each robot's URDF into joint-link modules and combines a recurrent encoder, graph-based Transformer, and per-joint decoder to generate actions across variable leg counts and joint configurations. To improve training stability, PolyMorph uses morphology-normalized rewards, adaptive robot sampling, and URDF-derived PD-gain initialization. We evaluate PolyMorph in simulation across bipedal, tripedal, and quadrupedal robots, including unseen morphology variants and platforms, and deploy the policy on real quadruped platforms operated in multiple configurations. The results show that a single PolyMorph policy generalizes across diverse robot morphologies without per-robot policy retraining.

\end{abstract}




\section{Introduction}
\FloatBarrier

\IEEEPARstart{I}{n} the field of robotics, significant progress has been made through various control methodologies, including Model Predictive Control (MPC)~\cite{youm2023imitating} and learning-based approaches such as imitation learning (IL)~\cite{peng2018deepmimic, zhao2023learning} and reinforcement learning (RL)~\cite{hwangbo2019learning, schulman2017proximal}. Among these, model-free RL has become a powerful tool for legged robot control because it directly optimizes policies from environment interactions without requiring explicit dynamics models. This makes RL particularly effective for complex and underactuated systems, enabling robust locomotion over challenging terrains such as rough slopes, stairs, and deformable terrain~\cite{miki2022learning, kumar2021rma}.

\begin{figure*}[t!]
\centering
\includegraphics[width=\linewidth]{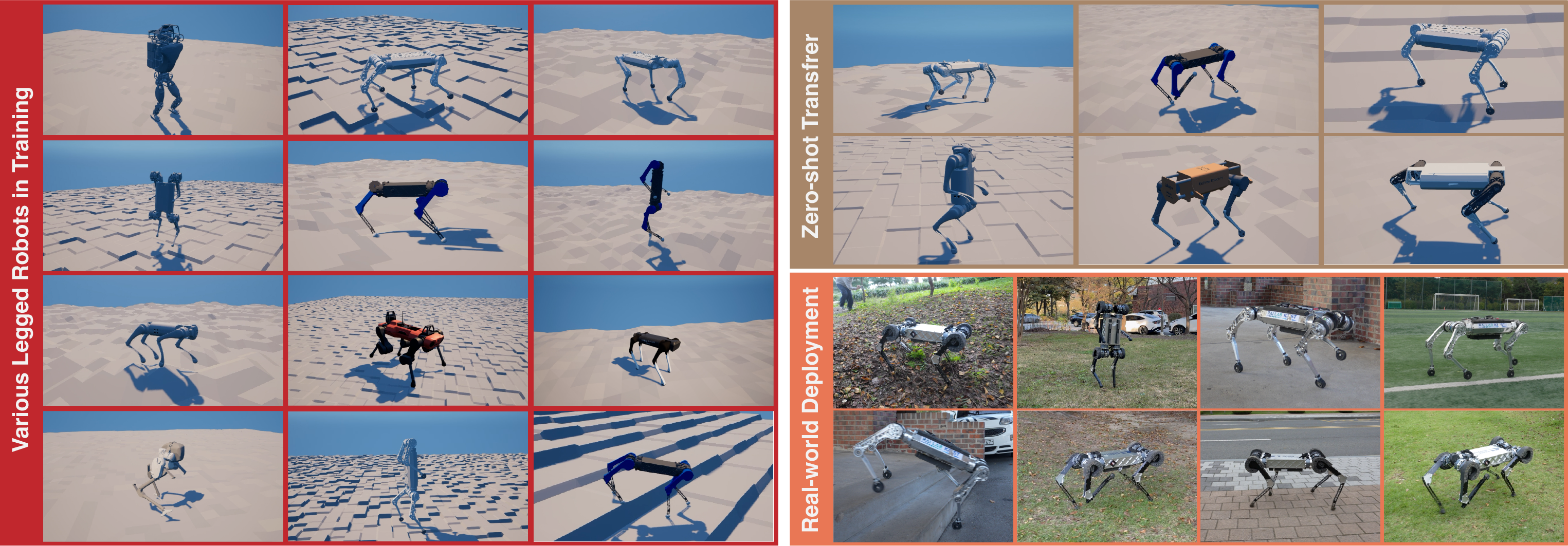}
\caption{\textbf{A single policy for cross-embodiment legged locomotion.}
PolyMorph is trained on diverse legged robots in simulation (left), zero-shot transfers to unseen morphology variants and platforms (top-right), and is deployed on real-world robots in multiple configurations (bottom-right).}
\label{fig:overall_description}
\vspace*{-0.25cm}
\end{figure*}

Most existing studies on legged locomotion focus on enabling a single platform to perform a specific task~\cite{kim2025high, ma2025learning}. However, policies trained for one robot often fail to transfer to another due to differences in morphology, including degrees of freedom, link masses, inertias, and kinematic configurations. Adapting a policy to a new robot typically requires redesigning the environment, observations, action space, and reward functions, which demands substantial engineering effort. This motivates the development of general controllers that can operate across diverse legged robots.

In this work, we address the following question: \textit{Can we train a single locomotion policy that operates across legged robots with diverse morphologies, including different numbers of legs and actuators, and deploy it on real hardware without per-robot retraining?} We argue that such a controller must not only observe the robot's current state but also reason about its morphology, and that achieving this in practice requires addressing training-level imbalances that arise when robots with very different reward magnitudes and learning difficulties are trained together.

To this end, we propose PolyMorph, a cross-embodiment locomotion framework with a URDF-conditioned policy interface. As shown in Fig.~\ref{fig:overall_description}, the same policy is deployed across diverse legged morphologies on various rough terrains in both simulation and the real world. Each robot is represented as a sequence of joint-link modules parsed from its URDF, and padding with valid-action masking allows the same policy to handle variable leg counts, actuator numbers, and joint configurations without redesigning the action interface. The policy combines a recurrent state encoder that summarizes proprioceptive history, a Transformer that conditions on URDF-derived graph structure, and per-joint decoder heads that generate actions only for actuated joints.

A unified locomotion policy across heterogeneous morphologies can shorten the robot design-test cycle by allowing new designs or modified configurations to be evaluated without retraining a separate controller for each platform, which is particularly useful for rapid robot design workflows~\cite{lee2024robotsketch, belmonte2022meta}. It also enables policy reuse across heterogeneous robot platforms and exposes the policy to broader morphology and dynamics variations.

Beyond designing a policy that can process robots with different kinematic structures, we observe that jointly optimizing across morphologies introduces training-level obstacles: reward magnitudes can scale with the number of joints and legs, biasing optimization toward certain morphologies; easy-to-learn robots can dominate uniform sampling and reduce the learning signal for harder morphologies such as bipeds; and manually tuned PD gains reintroduce per-platform engineering effort. Complementary to prior multi-embodiment methods that emphasize morphology-conditioned policy architectures, PolyMorph addresses these training-level obstacles through morphology-normalized rewards, adaptive robot sampling, URDF-derived PD-gain initialization, together with curriculum learning and domain randomization for robust training. The same policy parameters are shared across all robots, and only morphology-dependent inputs are derived from each robot's URDF.

In summary, the main contributions of this work are as follows:
\begin{itemize}
    \item We identify and address training-level obstacles in cross-embodiment optimization, including reward-scale mismatch, sampling imbalance between easy and hard robots, and per-platform gain tuning, through morphology-normalized rewards, adaptive robot sampling, and URDF-derived PD-gain initialization.
    
    \item We adapt joint-link tokenization, embodiment-masked attention, recurrent proprioceptive encoding, and per-joint action decoding into a URDF-conditioned policy interface that handles variable leg counts, actuator numbers, and kinematic structures with shared parameters.
    
    \item We validate PolyMorph on heterogeneous legged robots in simulation and demonstrate cross-morphology hardware deployment on two real quadruped platforms operated in quadrupedal, tripedal, and bipedal configurations, with additional ablations and comparisons against multi-embodiment baselines.
\end{itemize}

\section{Related Work}
\label{sect:related_work}

\subsection{Transformer architectures for robot control}

In legged robot control, Transformer-based policies have been used to process observation histories or sensory inputs for locomotion~\cite{radosavovic2024real}. Beyond such temporal use, Transformers also model relationships among input tokens through self-attention~\cite{vaswani2017attention}, making them suitable for control problems where multiple body parts must interact. This spatial formulation is particularly attractive for multi-embodiment locomotion, since variable-length token sequences naturally correspond to robots with different numbers of joints. However, standard Transformers do not explicitly encode the physical structure of the robot and must infer morphology-dependent relationships only from data, which can be sample-inefficient when training across heterogeneous robots. Body Transformer~\cite{sferrazza2024body} uses the robot's adjacency graph as an attention mask to inject structural bias for a fixed morphology. PolyMorph adopts this structural prior in a policy shared across heterogeneous morphologies, where URDF-derived morphology inputs and attention masks vary across embodiments and are combined with recurrent state encoding.

\subsection{Cross-embodied policy learning}

Cross-embodied policy learning aims to train a single policy that can operate across different robot embodiments. In legged robotics, Metamorph~\cite{gupta2022metamorph} uses a Transformer to control modular robots but is evaluated only in simulation. GenLoco~\cite{feng2023genloco} achieves real-world transfer through morphology randomization but focuses on quadrupeds. URMA~\cite{bohlinger2024one} learns multi-embodiment locomotion across quadrupedal and bipedal robots with an attention-based encoder--decoder. Graph neural networks~\cite{wang2018nervenet} encode robot morphology as graphs. GCNT~\cite{luo2025gcnt} further combines graph-based morphology encoding with Transformer communication to learn controllers across different robot structures. LocoFormer~\cite{liu2025locoformer} takes a different approach, combining large-scale procedural training with long-context adaptation to generalize locomotion across unseen embodiments. While these works have advanced policy architectures, the training-level aspects of jointly optimizing heterogeneous legged robots have received less attention. PolyMorph also targets them directly. Beyond locomotion, Chen et al.~\cite{chen2026cross} and Gupta et al.~\cite{gupta2026demonstrate} study cross-robot transfer from demonstrations through intention-aligned representations and kinematics-aware policy synthesis, respectively.

\begin{figure*}[t]
\centering
\includegraphics[width=\linewidth]{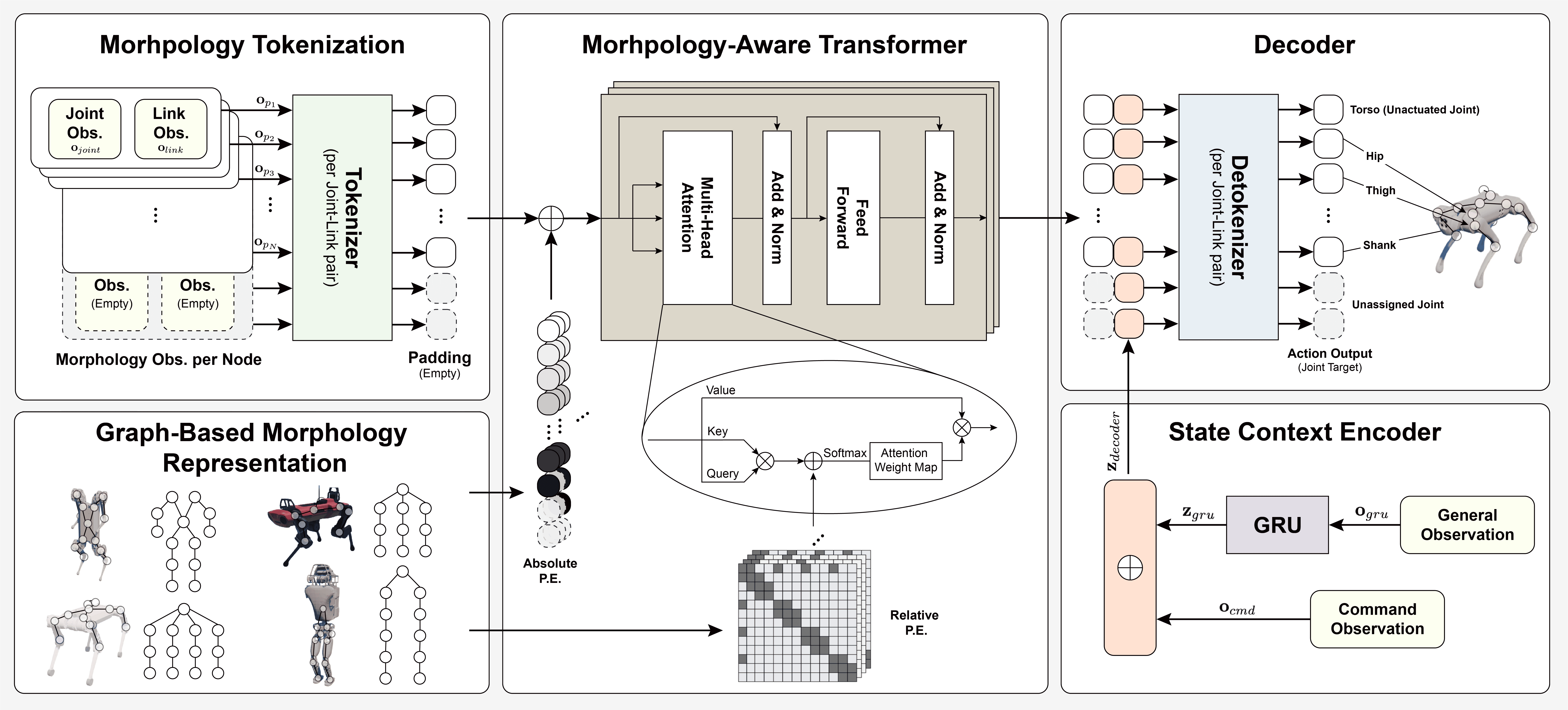}
\caption{\textbf{Overall architecture of PolyMorph.} PolyMorph consists of a GRU-based state context encoder, a morphology-aware Transformer that processes joint-link tokens using graph-based attention, and a decoder that generates actions for each actuated joint.}
\label{fig:learning_framework}
\vspace*{-0.25cm}
\end{figure*}

\section{Method}
\label{sect:framework}

We propose PolyMorph, a morphology-aware locomotion learning framework that controls legged robots with different morphologies using a single policy. For each robot $k$, we formulate the control problem as a Markov decision process (MDP) 
$\mathcal{M}^k=(\mathcal{S}^k,\mathcal{A}^k,T^k,R^k,\gamma)$. Given a set of $K$ robots with different morphologies, our objective is to learn one policy $\pi_\theta(a_t^k | s_t^k)$ that maximizes the average expected return:
\begin{equation}
\label{eq:rl_objective}
\max_{\theta} \frac{1}{K}\sum_{k=1}^{K}
\mathbb{E}_{\pi_\theta}
\left[
\sum_{t=0}^{H}\gamma^t r_t^k
\right].
\end{equation}
We optimize the policy using Proximal Policy Optimization (PPO)~\cite{schulman2017proximal}.

As shown in Fig.~\ref{fig:learning_framework}, PolyMorph consists of three components: a state context encoder, a morphology-aware Transformer, and a decoder. The state context encoder processes general proprioceptive observations over time. The morphology-aware Transformer represents each robot as a sequence of joint-link modules and extracts morphology-conditioned embeddings using graph-based attention. The decoder combines the state embedding, morphology embedding, and command input to generate target actions for each actuated joint.

\subsection{State Context Encoder}

The state context encoder uses a shared GRU to summarize proprioceptive history that is important for locomotion across robot morphologies. While each robot has different kinematic structure and actuator count, quantities such as body orientation, angular velocity, joint states, and previous target actions provide temporal information that is useful for balance, contact transitions, and recovery from disturbances. We therefore process these observations with a recurrent encoder and provide the resulting latent vector to the action decoder together with the morphology-conditioned Transformer features.

The actor observation is defined as
\begin{equation}
\mathbf{o}_{t,\mathrm{GRU}}
=
\left[
\mathbf{g}_t,\ \boldsymbol{\omega}_t,\ \mathbf{q}_t,\ 
\dot{\mathbf{q}}_t,\ \mathbf{q}^{des}_{t-1}
\right],
\end{equation}
where $\mathbf{g}_t$ is the roll-pitch vector of the body, $\boldsymbol{\omega}_t$ is the body angular velocity, $\mathbf{q}_t$ and $\dot{\mathbf{q}}_t$ are the joint positions and velocities, and $\mathbf{q}^{des}_{t-1}$ is the previous target joint position. These observations are obtained from the IMU and joint encoders. For robots with different numbers of joints, we use a configurable maximum number of actuated joints, $N_{\max}=15$ in the current experiments, and zero-pad the joint-related observations for robots with fewer joints. $N_{\max}$ can be increased for higher-DoF robot sets.

The GRU outputs a latent vector $\mathbf{z}_{\mathrm{gru}}$, which provides temporal context for action generation. We use an asymmetric actor-critic structure: the critic receives additional privileged information available in simulation, including body height, body linear velocity, and terrain height scan points, while the actor uses only onboard proprioceptive sensing at deployment.

\subsection{Morphology-Aware Transformer Network}

To handle robots with different numbers of joints and different kinematic structures, PolyMorph represents each robot as a sequence of joint-link modules. Each module consists of one joint and its attached child link, and is used as one Transformer token. The token observation is
\begin{equation}
\mathbf{o}_{p_i} =
\left[
\mathbf{o}_{link_i},\ \mathbf{o}_{joint_i}
\right],
\end{equation}
where $\mathbf{o}_{link_i}$ includes link mass, inertia, and center-of-mass position, while $\mathbf{o}_{joint_i}$ includes joint position, nominal joint angle, recent joint states, recent target actions, and PD gains. These quantities are either parsed from the URDF or updated from simulation and sensor observations.

The token sequence is ordered by traversing the robot's kinematic tree from the root body. Since robots may have different numbers of modules, unused token slots are filled with zero padding. Each token is projected through an independent linear layer and combined with an absolute positional embedding:
\begin{equation}
\label{eq:absolute_encoding}
\mathbf{x}_i =
\phi_i(\mathbf{o}_{p_i}; W_{e_i}) + \mathbf{w}_{pos,i},
\end{equation}
where $W_{e_i}$ is the token-specific projection parameter and $\mathbf{w}_{pos,i}$ encodes the position of the module in the ordered kinematic sequence. This absolute positional embedding provides the Transformer with information about the module order.

To inject morphology into the attention mechanism, we follow the embodiment-masked attention idea of Body Transformer~\cite{sferrazza2024body} and use the URDF-derived robot graph as a relative positional mask. Unlike a fixed-morphology setting, PolyMorph constructs joint-link tokens for each robot, handles different numbers of modules through padding, and later applies actions only to valid actuated joints. The self-attention operation is defined as
\begin{equation}
\label{eq:attention_equation}
\mathrm{Attention}(Q,K,V)
=
\mathrm{softmax}
\left(
\frac{QK^T}{\sqrt{d_k}} + r
\right)V,
\end{equation}
where $Q$, $K$, and $V$ are query, key, and value matrices, $d_k$ is the token dimension, and $r$ is the morphology-aware attention bias. We construct $r$ from the adjacency matrix $M$ \cite{sferrazza2024body} of the robot morphology graph:
\begin{equation}
\label{eq:relative_encoding}
r_{i,j} =
\begin{cases}
0, & \text{if } M_{i,j}=1, \\
-\infty, & \text{if } M_{i,j}=0.
\end{cases}
\end{equation}
Here, $M_{i,j}=1$ indicates that two modules are physically connected or allowed to attend to each other. This relative positional mask prevents attention to invalid structural connections and encourages the Transformer to reason over physically meaningful relationships. In this way, absolute positional embeddings encode the ordered kinematic sequence, while relative graph-based masking provides morphology-specific structural bias.

\subsection{Decoder Network}

The decoder generates actions by combining the morphology embedding, state embedding, and command input. For each module $i$, the decoder input is
\begin{equation}
\mathbf{z}_{i,\mathrm{dec}}
=
\left[
\mathbf{z}_{i,\mathrm{trans}},\ 
\mathbf{z}_{gru},\ 
\mathbf{o}_{cmd}
\right],
\end{equation}
where $\mathbf{z}_{i,\mathrm{trans}}$ is the Transformer output for module $i$, $\mathbf{z}_{gru}$ is the GRU output, and $\mathbf{o}_{cmd}$ is the desired velocity command.

For the actor, each module-specific decoder outputs the mean of a Gaussian action distribution:
\begin{equation}
\label{eq:action_generation}
\pi_\theta(a_{i}^{k} | s^{k}) = \mathcal{N}(\mu_{i}(\mathbf{z}_{i, decoder}; W_{d_i}),\;\Sigma),
\end{equation}
where $W_{d_i}$ is the independent decoder parameter for module $i$, and $\Sigma$ is a diagonal covariance matrix. Action samples from this distribution are applied only to the actuated joint parts corresponding to the specific ${k}$-th robot, excluding the first module (index 0), which represents the root element and is not actuated.

For the critic, the same decoder structure outputs a value estimate for each valid module. The final value is computed by averaging over the valid root and actuated modules:
\begin{equation}
V(s^{k}) = \frac{1}{N+1} \sum_{i=0}^{N} V_i,
\end{equation}
where $V_i$ is the value for the $i$-th module, $N$ is the number of actuated joints, and the summation starts from the root module at index 0. The resulting value function is used in PPO to compute the advantage and update the actor and critic.

\subsection{Training Techniques for Generalization}

When a single policy is trained to control robots with heterogeneous morphologies, three practical bottlenecks arise simultaneously: unequal learning difficulty across robots, morphology-dependent reward scales, and per-platform PD-gain tuning. PolyMorph integrates adaptive robot sampling, morphology-normalized rewards, and URDF-derived gain initialization to address these bottlenecks, respectively.

\textbf{Adaptive robot sampling.}
Uniformly sampling robots during training can under-train difficult morphologies such as bipeds. To balance learning, we increase the sampling probability of robots with lower rewards. The probability of sampling robot $i$ is computed as
\begin{equation}
\label{eq:adaptive_sampling}
P_i =
\frac{
\exp(\bar{r}-r_i)
}{
\sum_j \exp(\bar{r}-r_j)
},
\end{equation}
where $r_i$ is the reward of robot $i$ and $\bar{r}$ is the average reward over all robots, and the denominator ensures the probabilities sum to 1. To avoid overfitting to difficult robots, half of the environments are still sampled uniformly.

\textbf{Reward normalization.}
When learning a robot-agnostic policy, the reward should encourage motion generation and command tracking without favoring a specific robot morphology. However, if reward magnitudes depend strongly on the number of joints or legs, the policy can converge to a local optimum where only certain robot configurations achieve high rewards. Large reward-scale differences across robots can also destabilize PPO training and may even cause learning failure. To avoid this, we normalize morphology-dependent reward terms by the number of physical components that generate them. Joint-related penalties, such as joint torque, position, velocity, acceleration, and action smoothness, are divided by the number of actuated joints. Similarly, leg- or foot-related terms, such as airtime and foot clearance, are divided by the number of legs. This keeps reward scales consistent across morphologies and helps the policy learn robust behaviors without bias toward a particular robot configuration.

\textbf{Automatic PD-gain design.}
The policy outputs desired joint positions, which are converted to torques through a standard PD controller. To avoid robot-specific gain tuning in multi-embodiment training, we use a simple morphology-aware gain rule based on the URDF-derived mass matrix evaluated at a nominal posture.

We define a robot-level inertia scale from the URDF-derived mass matrix $\mathbf{M}(\mathbf{q}_0)$ at the nominal posture $\mathbf{q}_0$ as
\begin{equation}
\lambda = \sum_{i=1}^{3} M_{ii}(\mathbf{q}_{0}),
\end{equation}
which provides a morphology-dependent quantity for gain scaling.

The proportional gain is computed as
\begin{equation}
K_p = \alpha^2 \lambda,
\end{equation}
where $\alpha$ is a robot-agnostic scaling coefficient. We set $\alpha=0.8$ for all robots, which yields $K_p$ values in a practical range for the considered robots. The resulting scalar is applied uniformly to all actuators, i.e., $\mathbf{K}_{p}=K_p\mathbf{I}$. This rule replaces per-platform manual tuning with a single shared coefficient and a robot-specific inertial scale derived from the URDF.

For the derivative gain, we first compute a joint-wise damping value using the standard second-order damping relation:
\begin{equation}
K_{d,i} = 2\zeta\sqrt{M_{ii}(\mathbf{q}_{0})K_{p}},
\end{equation}
where $M_{ii}(\mathbf{q}_{0})$ is the diagonal joint-space inertia term of joint $i$ and $\zeta$ is the damping ratio. We set $\zeta=0.8$ to obtain a slightly underdamped response with small overshoot, balancing response speed and stability. Since $K_{d,i}$ can vary significantly between heavy proximal joints and light distal joints, directly using joint-wise values may introduce excessive damping variation. Therefore, we use the median value as a shared derivative gain:
\begin{equation}
K_{d}^{\ast} = \operatorname{median}\{K_{d,i}\}, \qquad \mathbf{K}_{d} = K_{d}^{\ast}\mathbf{I}.
\end{equation}
By grounding the gain scale in URDF-derived inertial quantities, the proposed rule provides morphology-aware PD gains that are computed once from $\mathbf{M}(\mathbf{q}_{0})$ and applied across different robots without manual per-platform tuning.

\textbf{Adjusting the Level of Difficulty}
We gradually increase task difficulty during training. A curriculum factor $c_k$ is updated as
\begin{equation}
c_{k+1}=c_k^\gamma,\qquad 0<\gamma<1,
\end{equation}
and environment parameters are interpolated between easy and hard settings:
\begin{equation}
p(k)=p_{\mathrm{easy}}+c_k(p_{\mathrm{hard}}-p_{\mathrm{easy}}).
\end{equation}
This schedule is applied to command ranges, terrain roughness, external disturbances, reward weights, and morphology perturbations. We also use a cosine-annealing learning-rate scheduler with warmup restarts to encourage exploration at the beginning of each curriculum stage and stable convergence afterward.

\textbf{Domain randomization.}
To improve sim-to-real transfer and robustness, we randomize both environmental and morphology-related parameters during training. These include observation noise, link and joint parameters, PD gains, control latency, friction, and external disturbances. The randomization is applied at the beginning of each episode or at every time step depending on the parameter.

\section{Results}
\label{sect:experiment_results}

Our experiments evaluate whether the overall PolyMorph framework enables stable training across heterogeneous robots, whether a single policy achieves locomotion across seen and unseen morphologies and transfers to real hardware, and whether PolyMorph improves performance relative to alternative architectures and multi-embodiment baselines.

\begin{figure}[t!] 
\centering
\includegraphics[width=\linewidth]{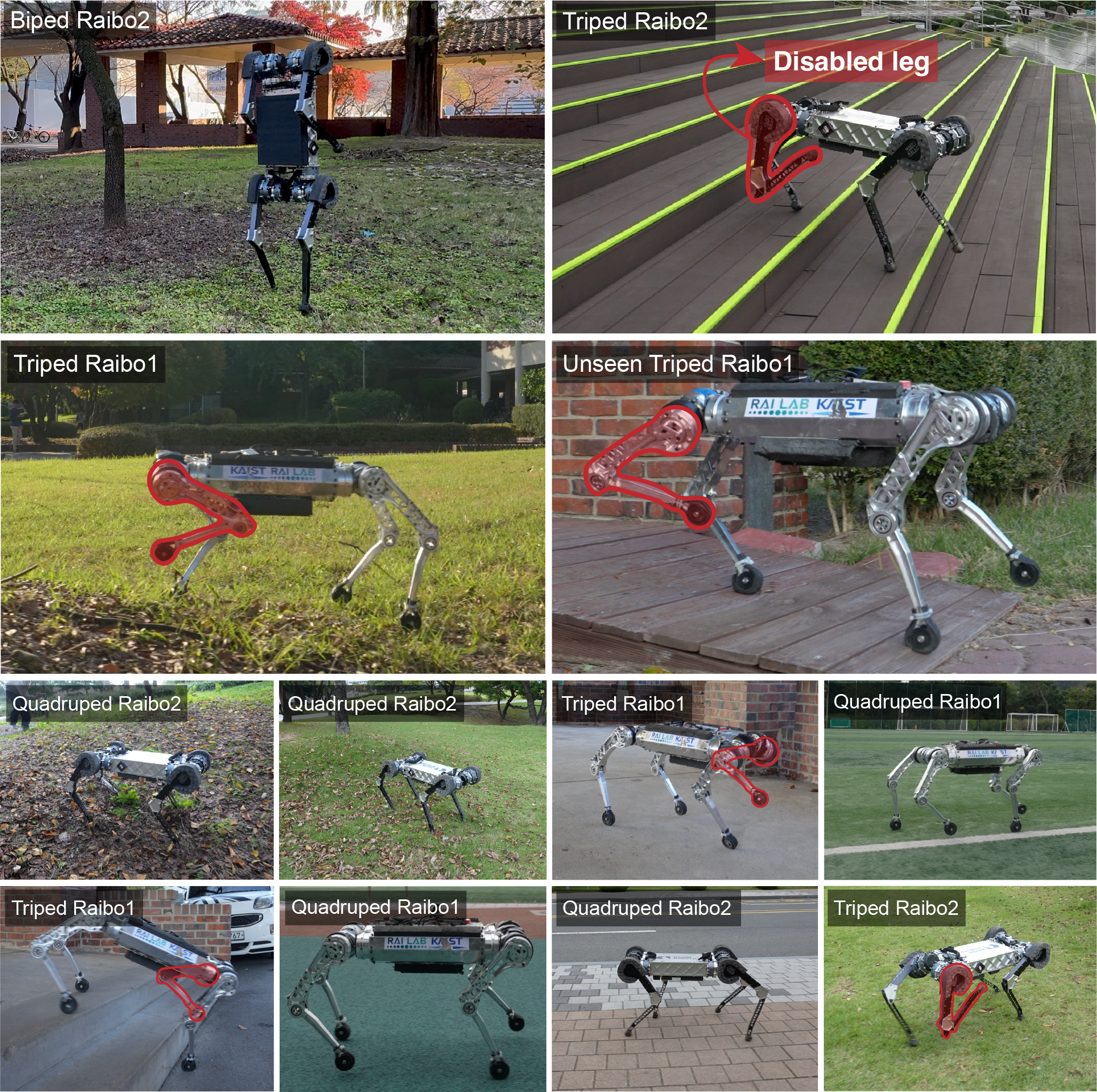}
\caption{A single trained controller is evaluated across diverse legged robot morphologies in simulation and deployed on Raibo1 and Raibo2 in bipedal, tripedal, and quadrupedal configurations. Both real-world deployment and unseen morphology configurations are tested on rough terrain.}
\label{fig:hardware_experiment}
\vspace*{-0.25cm}
\end{figure}

\subsection{Implementation Details}
\label{sec:experiment:implementation}
Training was conducted in RaiSim~\cite{hwangbo2018per} using 350 parallel environments for 20,000 iterations on an AMD EPYC 9654 CPU and an NVIDIA RTX 4090 GPU, requiring approximately 2.7 days. We trained three random seeds and evaluated each method on the same 100 episodes per robot. The deployed actor contains approximately 0.33 million parameters. For real-world deployment, the 100~Hz policy was exported to ONNX format and executed on the onboard computers of Raibo1 and Raibo2, with an average latency of 0.366~ms, as shown in Fig.~\ref{fig:hardware_experiment}.

\subsection{Framework Evaluation}
\label{sec:experiment:evaluation}

We trained a single locomotion policy across diverse legged robots, including Raibo1, Raibo2, Hound, Aliengo, Anymal-C, Atlas, Cassie, Go1, Laikago, Mini-Cheetah, and multiple morphology variants. The training set covers bipedal, tripedal, and quadrupedal configurations with different kinematic and dynamic properties, with masses ranging from 11\si{kg} to 135\si{kg}. Tripedal variants are generated by deleting one leg; FR, FL, HR, and HL denote the removed front-right, front-left, hind-right, and hind-left legs, respectively. The notation ``fix'' indicates that the corresponding leg is kept fixed instead of being removed.

\begin{table}[!t]
    \centering
    \caption{Comparison between automatic and manual PD gains}
    \begin{tabular}{|c|c|c|c|c|}
    \hline
        \multirow{2}{*}{\textbf{Robot}} & \multicolumn{2}{c|}{\textbf{Automatic method}} & \multicolumn{2}{c|}{\textbf{Manual reference}} \\ 
        \cline{2-5}
         & \textbf{$K_p$} & \textbf{$K_d$} & \textbf{$K_p$} & \textbf{$K_d$} \\
    \hline Raibo1              & 51 & 1.3 & 50  & 0.5 \\
    \hline Raibo2             & 77 & 2.7 & 80 & 1.0   \\
    \hline Hound              & 81 & 2.2 & 80  & 1.0  \\
    \hline Aliengo              & 40 & 1.2 & 40  & 2 \\
    \hline Anymal-C          & 100 & 2.0 & 80  & 2.0   \\
    \hline Atlas              & 260 & 3.6 & 300 & 4.0   \\
    \hline Go1                & 25 & 0.6 & 30 & 0.7 \\
    \hline Mini-Cheetah       & 21 & 0.8 & 20 & 0.5 \\
    \hline
    \end{tabular}
    \label{tab:pgain_dgain}
    \vspace*{-0.25cm}
\end{table}

\begin{table}[!t]
\centering
\caption{Reward-normalization ablation.}
\label{tab:reward_norm_ablation}
\setlength{\tabcolsep}{3.3pt}
\renewcommand{\arraystretch}{1.08}
\begin{tabular}{llcc}
\hline
Category & Metric & Full & w/o norm. \\
\hline
Early scale bias
& Joint-count corr. $|r|$ $\downarrow$ & \textbf{0.256} & 0.904 \\
\hline
\multirow{2}{*}{Scale imbalance}
& Joint-reward CV $\downarrow$         & \textbf{0.136} & 1.370 \\
& Foot-reward CV $\downarrow$          & \textbf{0.527} & 1.413 \\
\hline
\multirow{3}{*}{Outcome}
& Eval. return $\uparrow$              & \textbf{5.024} & 1.109 \\
& Avg. termination rate (\%) $\downarrow$ & \textbf{0.030} & 1.556 \\
& Termination-rate gap (pp) $\downarrow$       & \textbf{0.059} & 4.653 \\
\hline
\end{tabular}
\vspace{-0.2cm}
\end{table}

All morphologies are trained with the same policy, reward formulation, and training pipeline. Table~I shows that the URDF-derived PD gains are broadly comparable to manual reference gains, while requiring no robot-specific tuning. The same gain coefficients are also used without retuning for the unseen B2 and Solo12 evaluations in Table~IV, providing additional validation beyond the training platforms. All results use these automatic gains.

Table~\ref{tab:reward_norm_ablation} ablates morphology-normalized rewards by removing the actuator- and leg-count normalization factors while keeping all other settings unchanged. Since the training reward scales differ between the two variants, Eval. return is computed using the same evaluation reward function for both policies. Joint-count corr. measures the early-training correlation between joint-related reward scale and the number of actuated joints, while Joint-reward CV and Foot-reward CV measure final reward-scale variation across morphology groups. Without reward normalization, reward scales become more morphology-dependent, resulting in lower evaluation return and a larger termination-rate gap. These results support morphology-normalized rewards for balanced policy optimization across heterogeneous legged robots.

\begin{figure}[h!]
\centering
\includegraphics[width=\linewidth]{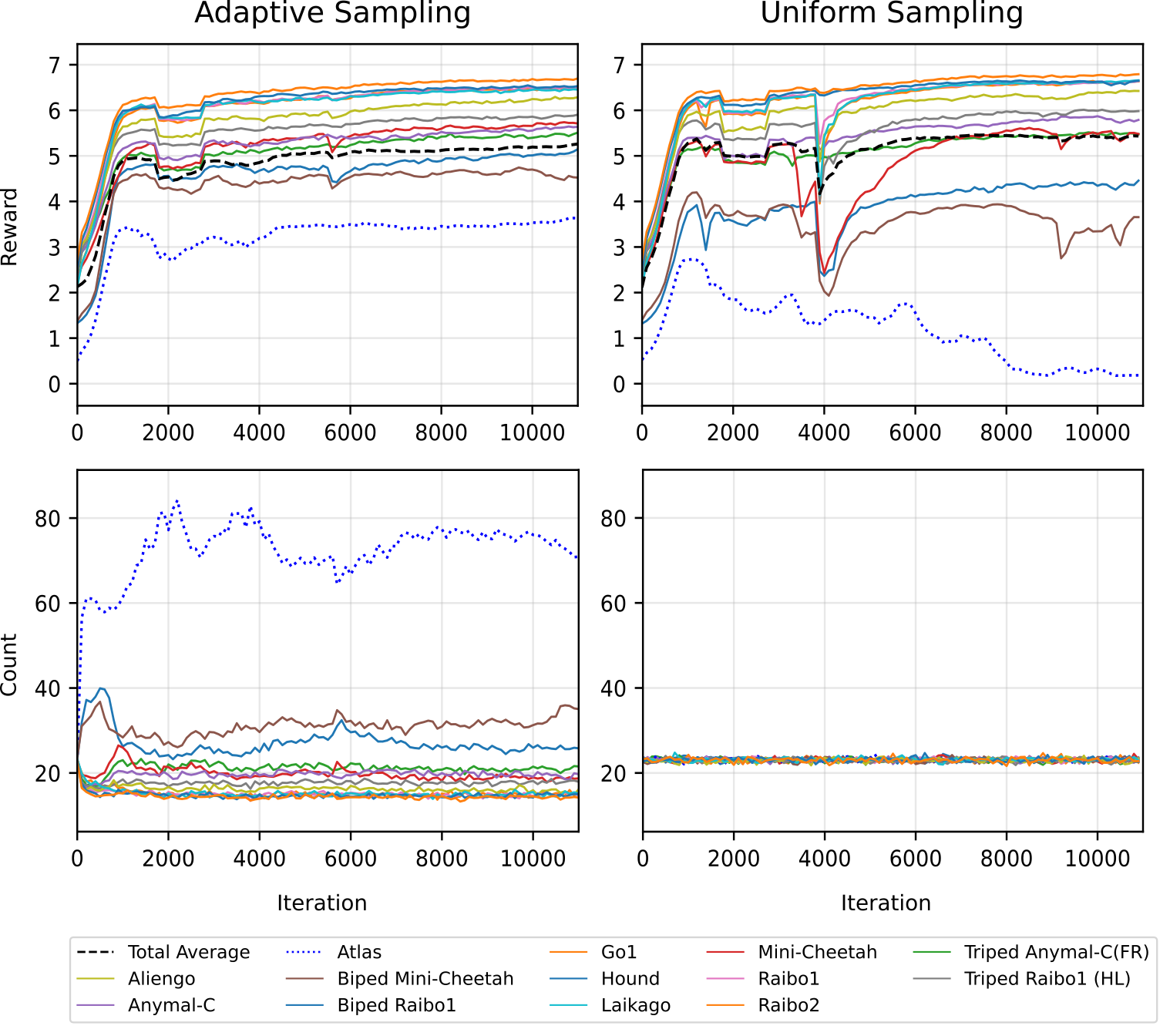}
\caption{Comparison of adaptive and uniform robot sampling strategies. The figures show the reward and the number of generations for each robot.}
\label{fig:adaptive_sampling}
\vspace*{-0.25cm}
\end{figure}

\begin{table}[!h]
\centering
\caption{Real-world blind-locomotion evaluation.}
\label{tab:hardware_results}
\begin{tabular}{lccc}
\hline
Configuration & Robot & Trials & Success \\
\hline
Quadruped & Raibo1, Raibo2 & 20 & 20 \\
Triped & Raibo1, Raibo2 & 20 & 20 \\
Biped & Raibo2 & 10 & 9 \\
Unseen triped & Raibo1 & 10 & 7 \\
\hline
\end{tabular}
\end{table}

Since easy-to-learn morphologies such as quadrupeds can dominate multi-robot training, harder morphologies such as bipeds may receive insufficient updates under uniform sampling. As shown in Fig.~\ref{fig:adaptive_sampling}, Atlas under uniform sampling fails to learn, with its reward decreasing as training progresses, while adaptive sampling allocates proportionally more environments to Atlas and restores stable convergence. This balances learning across diverse morphologies under limited computational resources.

We deployed the trained controller on Raibo1 and Raibo2 in quadrupedal, tripedal, and bipedal configurations, as shown in Fig.~\ref{fig:hardware_experiment}. In the tripedal experiments, the disabled leg remained attached with its joints locked, retaining mass and inertia absent from the leg-removed training morphologies. The same predefined 50~m outdoor route, comprising flat ground, slopes, steps, and naturally uneven terrain, was used for all configurations; success required completion without falling, emergency stop, or operator intervention. Table~\ref{tab:hardware_results} summarizes the results. Among the unsuccessful trials, failure modes differed between the unseen tripedal and bipedal configurations. For the unseen tripedal configuration, failures occurred when insufficient foot clearance caused the swing foot to catch on branches or raised terrain edges, while slip events were generally recovered. For the bipedal configuration, the failure resulted from severe balance loss, with point-foot contacts and a high center of mass limiting recovery. These hardware results demonstrate that a single PolyMorph policy can be deployed across multiple real robot configurations without per-configuration retraining.

\begin{figure}[t!]
\centering
\includegraphics[width=\linewidth]{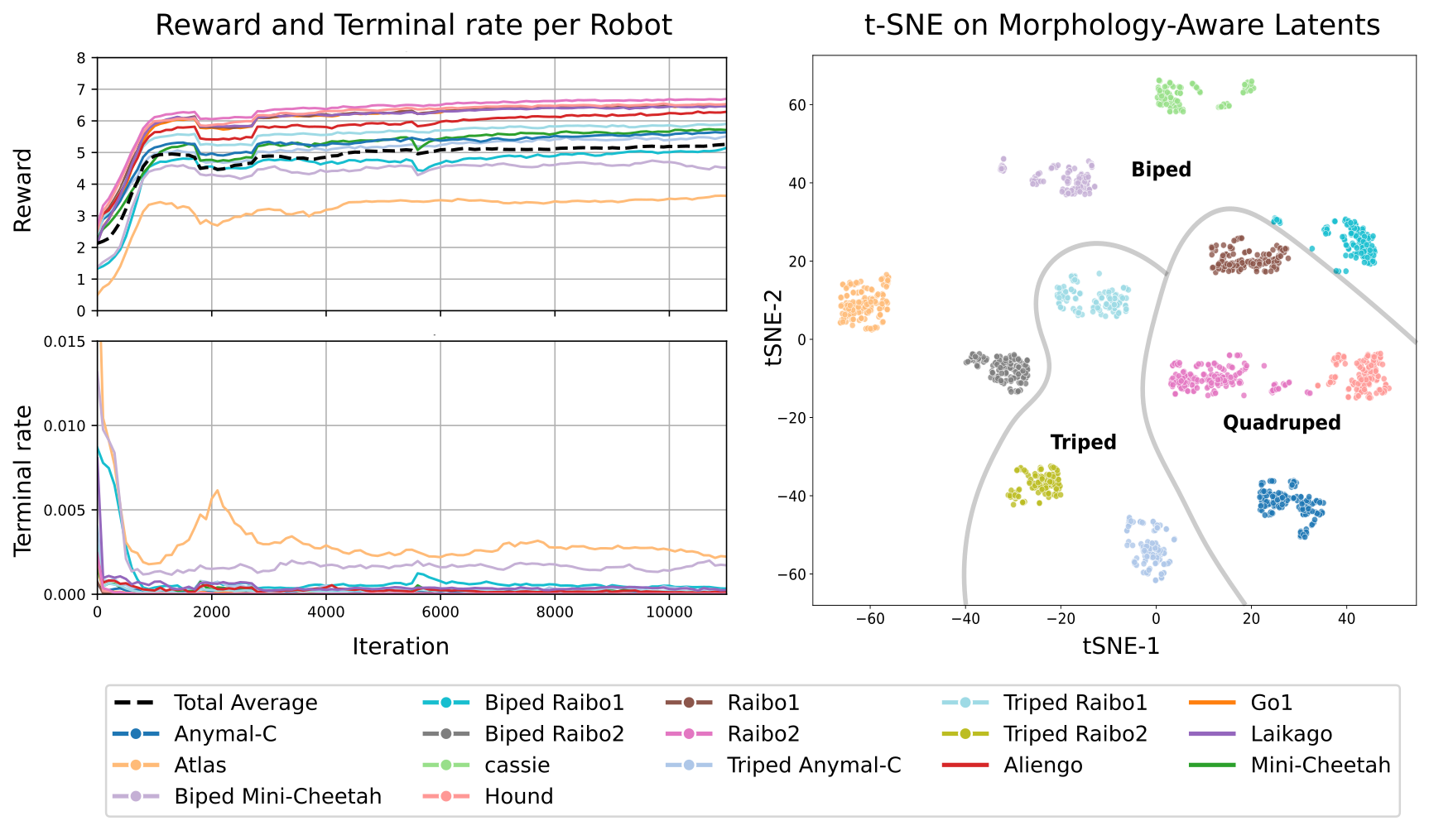}
\caption{The left panel shows rewards and termination rates. The right panel shows the t-SNE visualization of morphology-aware embeddings across bipedal, tripedal, and quadrupedal morphologies.}
\label{fig:Polymorph_mainResult}
\vspace*{-0.25cm}
\end{figure}

\begin{table}[h!]
\centering
\caption{COMMAND-TRACKING RMSE ACROSS DIFFERENT ROBOT MORPHOLOGIES}
\label{tab:command_tracking}
\begin{tabular}{|c|c|c|c|}
\hline
 & \textbf{Robot Model} & \textbf{$V_{xy}$ [m/s]} & \textbf{$\omega_z$ [rad/s]} \\
\hline
\multirow{10}{*}{Seen} & Raibo1              & 0.327(±0.112) & 0.176(±0.169) \\
\cline{2-4} & Triped Raibo1(HL)       & 0.142(±0.083) & 0.085(±0.083) \\
\cline{2-4} & Hound              & 0.168(±0.059) & 0.058(±0.057) \\
\cline{2-4} & Anymal-C           & 0.172(±0.117) & 0.092(±0.072) \\
\cline{2-4} & Atlas              & 0.246(±0.105) & 0.193(±0.185) \\
\cline{2-4} & Biped Mini-cheetah & 0.225(±0.078) & 0.154(±0.153) \\
\cline{2-4} & Cassie             & 0.204(±0.127) & 0.153(±0.152) \\
\cline{2-4} & Go1                & 0.152(±0.042)& 0.083(±0.080) \\
\hline
\multirow{5}{*}{Unseen} & Triped Hound(FL)   & 0.145(±0.073) & 0.072(±0.060) \\
\cline{2-4} & Biped Go1       & 0.323(±0.174) & 0.180(±0.163) \\
\cline{2-4} & Triped Anymal-C(HL)    & 0.270(±0.094) & 0.068(±0.066) \\
\cline{2-4} & B2    & 0.124(±0.068) & 0.168(±0.111) \\
\cline{2-4} & Solo12    & 0.254(±0.112) & 0.124(±0.121) \\
\hline
\end{tabular}
\end{table}

\subsection{Framework Analysis}
\label{sec:experiment:analyzation}

We analyze PolyMorph from two perspectives: cross-morphology control performance and morphology-aware representation learning. Using training curves and command-tracking results, we first evaluate whether a single policy maintains stable performance across robots with different leg counts, body configurations, and mechanical properties. We then use t-SNE embeddings to examine whether the graph-based Transformer captures meaningful morphology-dependent structure.

As shown in Fig.~\ref{fig:Polymorph_mainResult}, PolyMorph converges stably across diverse robot morphologies. Quadrupeds generally achieve higher rewards due to their greater static and dynamic stability, whereas bipeds obtain lower rewards because balancing with fewer contacts is more difficult on rough terrain. Nevertheless, the terminal rates decrease and remain low after training, indicating stable locomotion even for challenging morphologies.

For command-tracking evaluation, each robot was tested over multiple rough-terrain episodes with randomized velocity commands. The commands $v_x\sim\mathcal{U}(-1,2)$~m/s, $v_y\sim\mathcal{U}(-1,1)$~m/s, and $\omega_z\sim\mathcal{U}(-1,1)$~rad/s were sampled independently and held for 10~s. The first 2~s were excluded as a transient, and the RMSE was computed over the remaining 8~s and averaged across evaluation episodes. Table~\ref{tab:command_tracking} summarizes the command-tracking performance on rough terrain. PolyMorph achieves low RMSE for both seen and unseen robots, showing that the learned controller generalizes beyond the training morphologies. This generalization is enabled by representing each joint-link pair as an individual token and using graph-based attention to condition actions on the robot structure. Real-world experiments further confirm that the same policy can be deployed on different configurations while maintaining stable locomotion using only proprioceptive observations.

Among the unseen entries in Table~\ref{tab:command_tracking}, B2 and Solo12 evaluate quadrupedal extrapolation beyond the training set, which spans 11\si{kg} (Mini-Cheetah) to 52\si{kg} (Anymal-C). B2 (70\si{kg}) and Solo12 (2.5\si{kg}) represent heavy- and light-side extrapolation cases, respectively. PolyMorph achieves command-tracking accuracy on B2 comparable to seen platforms, while on Solo12, whose mass is less than one quarter of the lightest training quadruped and whose body scale and actuator characteristics differ substantially, it maintains command-following behavior with reduced tracking precision. Overall, PolyMorph transfers reliably across the quadrupedal training distribution and generalizes to morphologically similar out-of-distribution platforms, while extreme scale differences remain a direction for broader training coverage.

To examine whether the graph-based Transformer captures morphology information, we visualize the learned token embeddings using t-SNE. As shown in Fig.~\ref{fig:Polymorph_mainResult}, the embeddings form clusters according to the number of legs and robot body structure. Robots with similar size and kinematic properties, such as Raibo2 and Hound, appear close to each other, while different configurations of the same robot form nearby but distinct clusters. Unseen robots also cluster near morphologically similar robots, suggesting that PolyMorph encodes structural features rather than simply memorizing robot identities, and learns morphology-aware representations that support robust policy learning across diverse legged robots.

\begin{table*}[ht!]
\caption{Command-Tracking RMSE Comparison Across Different Networks}
\centering
\setlength{\tabcolsep}{5pt}
\renewcommand{\arraystretch}{1.15}
\footnotesize
\setlength{\arrayrulewidth}{0.6pt}
{
\scriptsize
\begin{tabular}{c l cc cc cc cc}
\hline
 & \textbf{Robot} &
\multicolumn{2}{c}{\textbf{Polymorph}} &
\multicolumn{2}{c}{\textbf{Polymorph(MLP)}} &
\multicolumn{2}{c}{\textbf{Metamorph}} &
\multicolumn{2}{c}{\textbf{URMA}} \\
\cline{3-10}
 &  &
 $V_{xy}$ [m/s] & $\omega_z$ [rad/s] &
 $V_{xy}$ [m/s] & $\omega_z$ [rad/s] &
 $V_{xy}$ [m/s] & $\omega_z$ [rad/s] &
 $V_{xy}$ [m/s] & $\omega_z$ [rad/s] \\
\hline

\multirow{4}{*}{Seen}
& Triped Raibo1(HL)       & 0.188(±0.100) & \textbf{0.098(±0.088)} & 0.173(±0.083) & 0.201(±0.106) & 1.136(±0.111) & 0.217(±0.160) & \textbf{0.172(±0.067)} & 0.147(±0.147) \\
& Atlas              & \textbf{0.479(±0.226)} & \textbf{0.204(±0.204)} & 0.516(±0.195) & 0.385(±0.236) & 1.260(±0.145) & 0.311(±0.194) & 0.744(±0.234) & 0.361(±0.286) \\
& Anymal-C            & 0.160(±0.091) & \textbf{0.082(±0.078)} & 0.229(±0.101) & 0.152(±0.110) & 0.995(±0.139) & 0.328(±0.201) & \textbf{0.107(±0.048)} & 0.104(±0.067) \\
& Biped Mini-cheetah  & \textbf{0.275(±0.103)} & \textbf{0.156(±0.144)} & 0.409(±0.106) & 0.240(±0.227) & 0.612(±0.113) & 0.617(±0.615) & \textbf{0.275(±0.077)} & 0.314(±0.280) \\
\hline

\multirow{4}{*}{Unseen}
& Triped Raibo2(HR-fixed) & \textbf{0.153(±0.103)} & \textbf{0.143(±0.143)} & - (fail) & - (fail) & 0.611(±0.331) & 0.630(±0.299) & 0.256(±0.083) & 0.176(±0.171) \\
& Triped Raibo1(HR) & 0.484(±0.100) & 0.195(±0.180) & \textbf{0.428(±0.076)} & \textbf{0.135(±0.126)} & - (fail) & - (fail) & - (fail) & - (fail) \\
& Hound              & \textbf{0.244(±0.075)} & \textbf{0.108(±0.102)} & 0.499(±0.094) & 0.191(±0.092) & 1.298(±0.058) & 0.292(±0.159) & - (fail) & - (fail) \\
& Laikago            & \textbf{0.090(±0.051)} & \textbf{0.160(±0.087)} & 0.377(±0.059) & 0.255(±0.147) & 0.840(±0.231) & 0.231(±0.174) & 0.131(±0.090) & 0.265(±0.209) \\
\hline
\end{tabular}
}
\label{tab:cmd_tracking_ModelComparison}
\end{table*}

\begin{figure}[ht!]
\centering
\includegraphics[width=\linewidth]{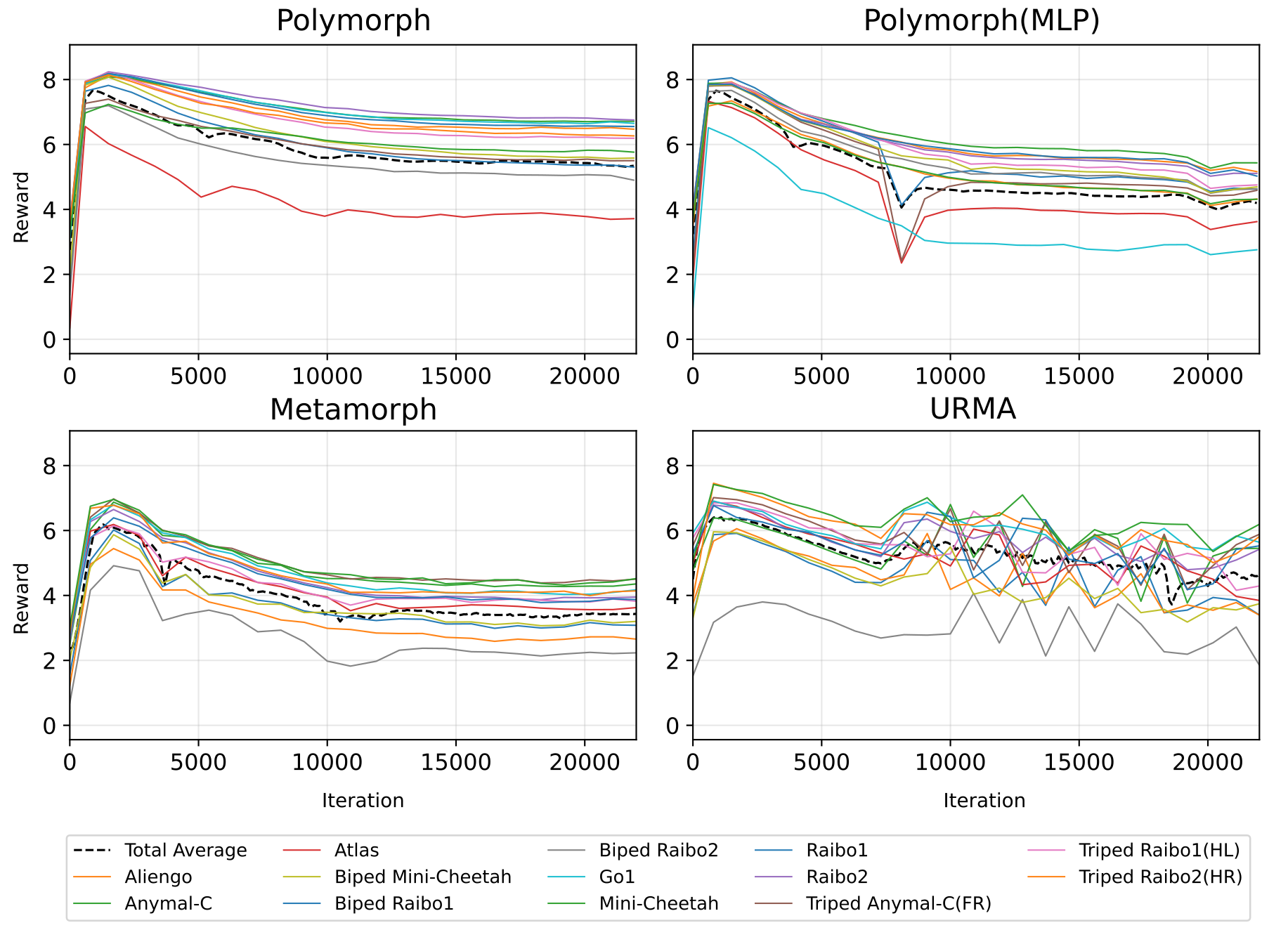}
\caption{Comparison of different networks. The figures show the total reward and per-robot reward for each network.}
\label{fig:compareNetwork}
\vspace*{-0.25cm}
\end{figure}

\subsection{Ablation and Baseline Comparison} 
\label{sec:experiment:ablation} 
We evaluate PolyMorph from two perspectives. First, we examine structured token-wise processing by comparing PolyMorph with an MLP variant that flattens the joint-link features into a single vector instead of processing them as separate Transformer tokens, while keeping the same training pipeline. Second, we compare PolyMorph with representative multi-embodiment baselines, Metamorph~\cite{gupta2022metamorph} and URMA~\cite{bohlinger2024one}, under comparable training budgets. For controlled comparison, we preserve the method-specific observations, morphology representations, and network architectures of Metamorph and URMA following their original publications and official implementations, while training and evaluating all methods under identical training conditions, budget, and evaluation protocol. Each method is trained with 3 random seeds and evaluated under the protocol described in Sec.~\ref{sec:experiment:implementation}. In Table~\ref{tab:cmd_tracking_ModelComparison}, "Fail" indicates termination during evaluation (e.g., falling).

As shown in Fig.~\ref{fig:compareNetwork}, PolyMorph achieves consistently higher total reward than the baselines and maintains balanced per-robot performance across diverse morphologies. The modular joint-link representation and graph-based Transformer attention allow the policy to decompose robot morphology into structured components and model their interactions, leading to more stable policy learning across heterogeneous robots. 

The PolyMorph-MLP variant degrades in performance as the number and diversity of robots increase. This indicates that the graph-based Transformer is important for interpreting morphology and generalizing to new structures. Metamorph and URMA both rely on attention mechanisms but lack explicit morphology-aware structural bias and recurrent temporal processing of proprioceptive observations. As a result, Metamorph struggles to track commands and fails on some seen and unseen robots, while URMA remains competitive on some training robots but becomes unstable during later curriculum stages and fails on some unseen robots. These results suggest that attention-based architectures without explicit morphology-aware bias may not be sufficient to fit heterogeneous morphologies jointly. Together with the PolyMorph-MLP ablation, the comparison suggests that both morphology-aware graph attention and the training techniques contribute to PolyMorph's training stability, command tracking, and generalization to unseen morphologies.

\section{Conclusion and Future Work}
\label{sect:conclusion_future_work}

In this work, we presented PolyMorph, a morphology-conditioned locomotion framework for controlling legged robots with diverse morphologies using a single policy. PolyMorph represents each robot as a URDF-derived sequence of joint-link modules and combines a recurrent state encoder, a graph-based Transformer, and per-joint decoder heads to generate actions across heterogeneous morphologies. Combined with morphology-normalized rewards, adaptive robot sampling, URDF-derived PD-gain initialization, curriculum learning, and domain randomization, the framework improves training stability across robots with different leg counts, configurations, and mechanical properties. Simulation and real-world experiments demonstrate robust command tracking on rough terrain across bipedal, tripedal, and quadrupedal robots, including unseen configurations of training platforms and unseen quadrupedal platforms.

The main limitation of the current framework is that generalization is reliable only within the morphology range covered by training and degrades under strong extrapolation in topology. Future work will focus on overcoming this limitation through robot data augmentation, such as procedurally perturbing URDFs and synthesizing morphology variants along the topology axis, so that the network is exposed to a broader range of robots during training and can extrapolate more reliably to platforms substantially different from the training set.



\ifCLASSOPTIONcaptionsoff
  \newpage
\fi


\typeout{}
\bibliographystyle{bibtex/IEEEtran}
\bibliography{bibtex/IEEEabrv,bibtex/library}

@article{hwangbo2019learning,
  title={Learning agile and dynamic motor skills for legged robots},
  author={Hwangbo, Jemin and Lee, Joonho and Dosovitskiy, Alexey and Bellicoso, Dario and Tsounis, Vassilios and Koltun, Vladlen and Hutter, Marco},
  journal={Science Robotics},
  volume={4},
  number={26},
  pages={eaau5872},
  year={2019},
  publisher={American Association for the Advancement of Science}
}

@article{miki2022learning,
  title={Learning robust perceptive locomotion for quadrupedal robots in the wild},
  author={Miki, Takahiro and Lee, Joonho and Hwangbo, Jemin and Wellhausen, Lorenz and Koltun, Vladlen and Hutter, Marco},
  journal={Science robotics},
  volume={7},
  number={62},
  pages={eabk2822},
  year={2022},
  publisher={American Association for the Advancement of Science}
}

@article{youm2023imitating,
  title={Imitating and finetuning model predictive control for robust and symmetric quadrupedal locomotion},
  author={Youm, Donghoon and Jung, Hyunyoung and Kim, Hyeongjun and Hwangbo, Jemin and Park, Hae-Won and Ha, Sehoon},
  journal={IEEE Robotics and Automation Letters},
  year={2023},
  publisher={IEEE}
}

@article{kumar2021rma,
  title={Rma: Rapid motor adaptation for legged robots},
  author={Kumar, Ashish and Fu, Zipeng and Pathak, Deepak and Malik, Jitendra},
  journal={arXiv preprint arXiv:2107.04034},
  year={2021}
}

@article{peng2018deepmimic,
  title={Deepmimic: Example-guided deep reinforcement learning of physics-based character skills},
  author={Peng, Xue Bin and Abbeel, Pieter and Levine, Sergey and Van de Panne, Michiel},
  journal={ACM Transactions On Graphics (TOG)},
  volume={37},
  number={4},
  pages={1--14},
  year={2018},
  publisher={ACM New York, NY, USA}
}

@article{schulman2017proximal,
  title={Proximal policy optimization algorithms},
  author={Schulman, John and Wolski, Filip and Dhariwal, Prafulla and Radford, Alec and Klimov, Oleg},
  journal={arXiv preprint arXiv:1707.06347},
  year={2017}
}

@article{gupta2022metamorph,
  title={Metamorph: Learning universal controllers with transformers},
  author={Gupta, Agrim and Fan, Linxi and Ganguli, Surya and Fei-Fei, Li},
  journal={arXiv preprint arXiv:2203.11931},
  year={2022}
}

@article{vaswani2017attention,
  title={Attention is all you need},
  author={Vaswani, A},
  journal={Advances in Neural Information Processing Systems},
  year={2017}
}

@article{radosavovic2024real,
  title={Real-world humanoid locomotion with reinforcement learning},
  author={Radosavovic, Ilija and Xiao, Tete and Zhang, Bike and Darrell, Trevor and Malik, Jitendra and Sreenath, Koushil},
  journal={Science Robotics},
  volume={9},
  number={89},
  pages={eadi9579},
  year={2024},
  publisher={American Association for the Advancement of Science}
}

@inproceedings{wang2018nervenet,
  title={Nervenet: Learning structured policy with graph neural networks},
  author={Wang, Tingwu and Liao, Renjie and Ba, Jimmy and Fidler, Sanja},
  booktitle={International conference on learning representations},
  year={2018}
}

@article{sferrazza2024body,
  title={Body Transformer: Leveraging Robot Embodiment for Policy Learning},
  author={Sferrazza, Carmelo and Huang, Dun-Ming and Liu, Fangchen and Lee, Jongmin and Abbeel, Pieter},
  journal={arXiv preprint arXiv:2408.06316},
  year={2024}
}

@article{bohlinger2024one,
  title={One policy to run them all: an end-to-end learning approach to multi-embodiment locomotion},
  author={Bohlinger, Nico and Czechmanowski, Grzegorz and Krupka, Maciej and Kicki, Piotr and Walas, Krzysztof and Peters, Jan and Tateo, Davide},
  journal={arXiv preprint arXiv:2409.06366},
  year={2024}
}

@article{hwangbo2018per,
  title={Per-contact iteration method for solving contact dynamics},
  author={Hwangbo, Jemin and Lee, Joonho and Hutter, Marco},
  journal={IEEE Robotics and Automation Letters},
  volume={3},
  number={2},
  pages={895--902},
  year={2018},
  publisher={IEEE}
}

@inproceedings{feng2023genloco,
  title={Genloco: Generalized locomotion controllers for quadrupedal robots},
  author={Feng, Gilbert and Zhang, Hongbo and Li, Zhongyu and Peng, Xue Bin and Basireddy, Bhuvan and Yue, Linzhu and Song, Zhitao and Yang, Lizhi and Liu, Yunhui and Sreenath, Koushil and others},
  booktitle={Conference on Robot Learning},
  pages={1893--1903},
  year={2023},
  organization={PMLR}
}

@article{kim2025high,
  title={High-speed control and navigation for quadrupedal robots on complex and discrete terrain},
  author={Kim, Hyeongjun and Oh, Hyunsik and Park, Jeongsoo and Kim, Yunho and Youm, Donghoon and Jung, Moonkyu and Lee, Minho and Hwangbo, Jemin},
  journal={Science Robotics},
  volume={10},
  number={102},
  pages={eads6192},
  year={2025},
  publisher={American Association for the Advancement of Science}
}

@article{zhao2023learning,
  title={Learning fine-grained bimanual manipulation with low-cost hardware},
  author={Zhao, Tony Z and Kumar, Vikash and Levine, Sergey and Finn, Chelsea},
  journal={arXiv preprint arXiv:2304.13705},
  year={2023}
}

@incollection{lee2024robotsketch,
  title={RobotSketch: an interactive showcase of superfast design of legged robots},
  author={Lee, Joon Hyub and Oh, Hyunsik and Yoon, Junwoo and Lee, Seung-Jun and Jin, Taegyu and Hwangbo, Jemin and Bae, Seok-Hyung},
  booktitle={ACM SIGGRAPH 2024 Emerging Technologies},
  pages={1--2},
  year={2024}
}

@article{ma2025learning,
  title={Learning coordinated badminton skills for legged manipulators},
  author={Ma, Yuntao and Cramariuc, Andrei and Farshidian, Farbod and Hutter, Marco},
  journal={Science Robotics},
  volume={10},
  number={102},
  pages={eadu3922},
  year={2025},
  publisher={American Association for the Advancement of Science}
}

@article{belmonte2022meta,
  title={Meta reinforcement learning for optimal design of legged robots},
  author={Belmonte-Baeza, {\'A}lvaro and Lee, Joonho and Valsecchi, Giorgio and Hutter, Marco},
  journal={IEEE Robotics and Automation Letters},
  volume={7},
  number={4},
  pages={12134--12141},
  year={2022},
  publisher={IEEE}
}

@article{chen2026cross,
  title={Cross-robot behavior adaptation through intention alignment},
  author={Chen, Xi and Gao, Yuan and Liu, Hangxin and Yang, Fangkai and Ghadirzadeh, Ali and Yang, Jun and Liang, Bin and Zhang, Chongjie and Lam, Tin Lun and Zhu, Song-Chun},
  journal={Science Robotics},
  volume={11},
  number={112},
  pages={eadv2250},
  year={2026},
  publisher={American Association for the Advancement of Science}
}

@article{gupta2026demonstrate,
  title={Demonstrate once, execute on many: Kinematic intelligence for cross-robot skill transfer},
  author={Gupta, Sthithpragya and Salunkhe, Durgesh Haribhau and Billard, Aude},
  journal={Science Robotics},
  volume={11},
  number={113},
  pages={eaea1995},
  year={2026},
  publisher={American Association for the Advancement of Science}
}

@article{liu2025locoformer,
  title={Locoformer: Generalist locomotion via long-context adaptation},
  author={Liu, Min and Pathak, Deepak and Agarwal, Ananye},
  journal={arXiv preprint arXiv:2509.23745},
  year={2025}
}

@article{luo2025gcnt,
  title={Gcnt: Graph-based transformer policies for morphology-agnostic reinforcement learning},
  author={Luo, Yingbo and Yao, Meibao and Xiao, Xueming},
  journal={arXiv preprint arXiv:2505.15211},
  year={2025}
}

\end{document}